\documentclass[conference]{IEEEtran}
\IEEEoverridecommandlockouts

\usepackage{cite}
\usepackage{amsmath,amssymb,amsfonts}
\usepackage{algorithmic}
\usepackage{graphicx}
\usepackage{textcomp}
\usepackage{xcolor}
\usepackage{booktabs}
\usepackage{multirow}
\usepackage{makecell}
\usepackage{threeparttable}
\usepackage{tikz}
\usetikzlibrary{shapes,arrows,positioning,calc,fit,backgrounds,decorations.pathreplacing}

\def\BibTeX{{\rm B\kern-.05em{\sc i\kern-.025em b}\kern-.08em
    T\kern-.1667em\lower.7ex\hbox{E}\kern-.125emX}}

\begin{document}

\title{Cross-Modal Generative Framework for Signal Translation from Fetal–Maternal Electrocardiograms to Fetal Doppler Waveforms }

\author{\IEEEauthorblockN{Tongli~Su\(^{1}\), Alireza~Rafiei\(^{1}\), Marly~van~Assen\(^{1}\), Reza~Sameni\(^{1,2}\),\\ Gari~D.~Clifford\(^{1,2}\), Faezeh~Marzbanrad\(^{3}\), Nasim~Katebi\(^{1}\)}
\IEEEauthorblockA{\(^{1}\)Emory University, Atlanta, GA, USA, \(^{2}\)Georgia Institute of Technology, Atlanta, GA, USA\\\(^{3}\)Monash University, Clayton, VIC, Australia\\
\thanks{{\scriptsize This work was supported in part by the NIH (R01HD110480) and Google.org AI for the Global Goals Impact Challenge. Email: \{ken.su\}@emory.edu}}}}

\maketitle
\begin{abstract}
Fetal electrocardiogram (fECG) and Doppler ultrasound provide complementary views of fetal cardiovascular function: fECG captures electrical activity while Doppler reflects mechanical hemodynamics shaped by factors such as placental resistance and vascular compliance. Understanding the recoverable and unrecoverable Doppler components through reconstruction from fECG offers insight into the relative contributions of electrical versus mechanical factors in fetal circulation, thereby informing clinical decisions. In addition, clinical evidence of maternal--fetal cardiac coupling suggests that maternal cardiovascular dynamics may also inform fetal hemodynamics. To computationally model these relationships, we propose a cross-modal generative framework combining dilated convolutions with cross-modal attention to selectively incorporate maternal ECG and self-attention to capture long-range temporal dependencies. Trained on 885 synchronized fetal/maternal ECG and Doppler envelope segments from 39 pregnancies, our model synthesizes Doppler envelopes with power spectral density mean squared error (PSD MSE) of $49.9\pm15.8$\,dB$^2$ (51\% lower than two-channel baseline) and heart-rate error of $4.71\pm0.77$\,bpm (1.5\% better than baseline; negligible relative to the 110--160\,bpm physiological range). Cross-modal attention yields 39\% PSD MSE reduction over na\"ive dual-channel concatenation, quantifying the contribution of maternal--fetal coupling. Our proposed framework advances computational modeling of the maternal--fetal cardiovascular system by enabling the synthesis of Doppler envelopes from dual-lead ECG. By analysis of both recoverable and residual Doppler components, this approach enables quantification of the purely mechanical contributions to Doppler waveforms---those not recoverable from electrical recordings---ultimately facilitating a more comprehensive fetal assessment. 
\end{abstract}

\section{Introduction}

Fetal electrocardiogram (fECG) and Doppler ultrasound provide complementary views of fetal cardiovascular function. The fECG captures the electrical activity of the heart, primarily heart rate and rhythm, while Doppler reflects mechanical hemodynamics including blood flow velocity patterns \cite{behar2014fecg}. Clinical indices derived from Doppler waveforms characterize diastolic function and overall cardiac performance \cite{hernandez2012evaluation, tei2004fetal}. However, Doppler acquisition requires expensive equipment and skilled sonographers, limiting accessibility in resource-constrained settings \cite{behar2014fecg}. In contrast, fECG can be collected widely and inexpensively but lacks direct measurements of mechanical function.
Rather than treating these modalities as interchangeable, we recognize that fECG and Doppler capture fundamentally different aspects of fetal cardiac physiology. Excitation-contraction coupling links electrical depolarization to mechanical contraction: each R-peak marks systolic onset after an electromechanical delay of roughly 60--80\,ms \cite{kahler2002fetal, bers2002cardiac}. Correlations between ECG features and contraction vigor suggest that fECG encodes partial information about mechanical events \cite{mensahbrown2010electromechanical}. However, Doppler waveforms are also shaped by purely mechanical factors, including ventricular compliance, valve function, and preload conditions, which have no direct electrical signature \cite{hernandez2012evaluation}. Understanding which Doppler-based signatures can be recovered from fECG, and which cannot, offers insight into the relative contributions of electrical versus mechanical factors in fetal cardiac function.
Clinical evidence further suggests that maternal cardiovascular dynamics may inform fetal hemodynamics through maternal--fetal cardiac coupling (MFCC). Van Leeuwen \textit{et al.} demonstrated phase synchronization between maternal and fetal heartbeats \cite{vanleeuwenpnas2009}, while transfer entropy analysis has quantified significant bidirectional information transfer that strengthens with gestational age \cite{marzbanrad2015transfer}. A scoping review reported that 91.3\% of studies investigating MFCC detected at least occasional coupling, where the failure of traditional linear methods to identify this interaction suggests a complex and non-linear relationship~\cite{nichting2023maternal}. This coupling likely occurs through hemodynamic interactions within the circulatory system, vibroacoustic transmission or autonomic links, though the underlying physiological pathways have not been definitively verified \cite{nichting2023maternal}. These findings motivate including maternal ECG as an auxiliary input for modeling fetal hemodynamics.

\subsection{Computational Modeling of Maternal--Fetal Cardiac Relationships}

To computationally model relationships between fECG, maternal ECG, and fetal Doppler, we require an architecture capable of selectively integrating information from multiple sources. We hypothesize that maternal influences on fetal hemodynamics are conditional and time-varying, occurring intermittently rather than uniformly. Simply concatenating maternal and fetal signals treats all maternal information as equally relevant, failing to exploit this selective coupling.

Cross-modal attention addresses this limitation by treating fetal features as queries and maternal features as keys/values, enabling the model to emphasize maternal ECG segments temporally correlated with fetal blood flow while suppressing unrelated noise \cite{vaswani2017attention}. Self-attention then captures long-range dependencies within the fetal signal, modeling how early cardiac events influence later phases. Together, these mechanisms enable investigation of both the recoverable Doppler components, reflecting excitation-contraction coupling and maternal--fetal interactions, and residual components that may reveal purely mechanical influences.

\subsection{Contributions}

Our main contributions are:
\begin{enumerate}
    \item We propose a novel cross-modal generative framework combining dilated convolutions with cross-modal and self-attention to model the relationship between fetal/maternal ECG and fetal Doppler waveforms. This approach enables the computational investigation of maternal--fetal cardiac coupling and the decomposition of Doppler components into electrically recoverable and residual mechanical components.
    \item We demonstrate that na\"ive inclusion of maternal ECG provides no benefit and can degrade performance; only when the model learns to selectively attend to maternal signals does the coupling information improve Doppler reconstruction, quantifying the contribution of maternal--fetal cardiac interactions.
    \item We introduce a composite loss function for physiological waveform reconstruction by balancing pointwise error, derivative error, and correlation to enforce both amplitude accuracy and morphological fidelity.
\end{enumerate}

\begin{figure*}[t]
\centering
\includegraphics[trim={0 1.0cm 0 0},clip,width=0.8\textwidth]{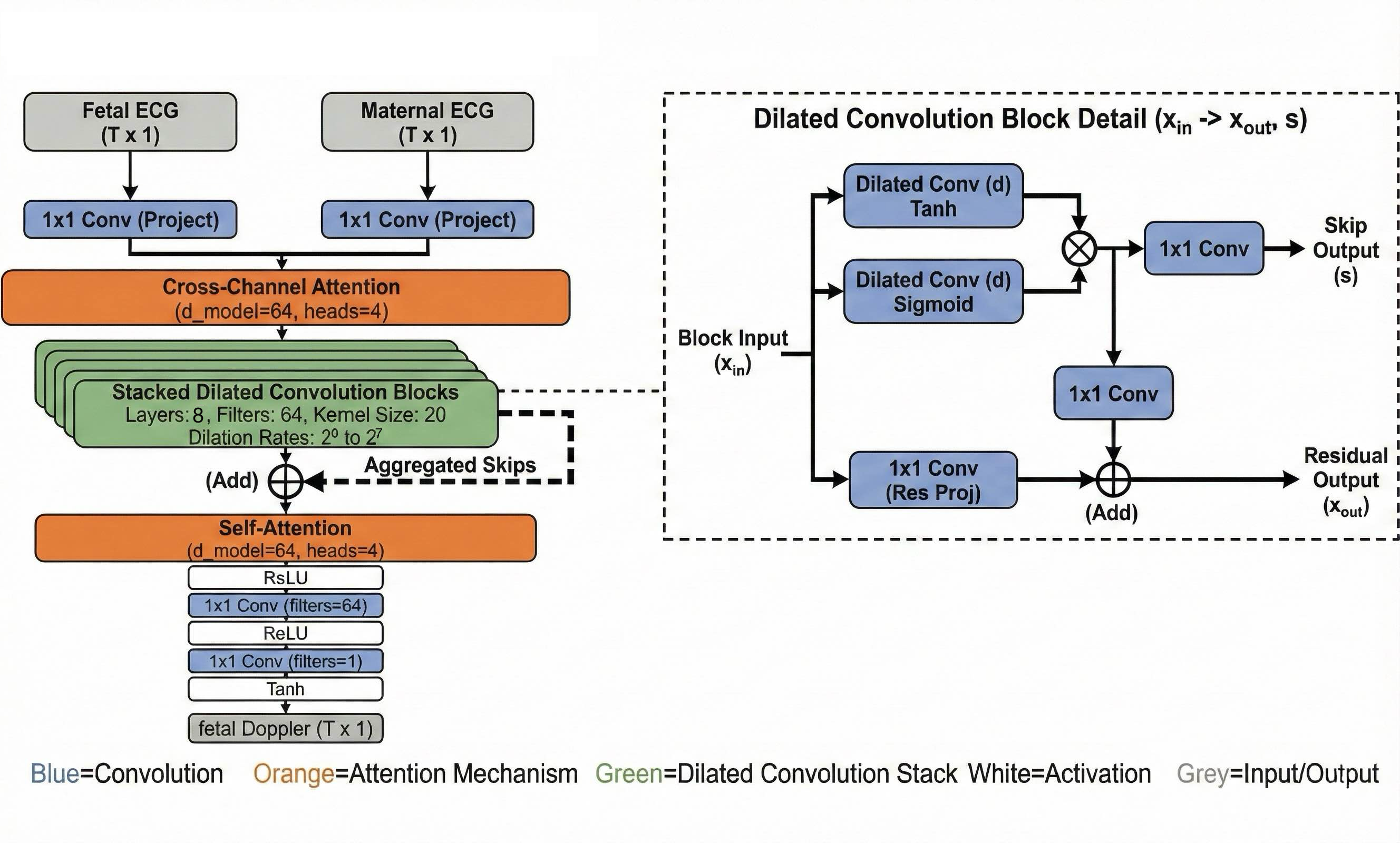}
\caption{Overview of the proposed cross-modal generative model. Dilated convolution layers extract features from fetal and maternal ECG, cross-modal attention selectively fuses maternal information based on fetal queries, and self-attention captures long-range dependencies to generate Doppler velocity envelopes.}
\label{fig:model_framework}
\end{figure*}

\section{Background and Related Work}

\subsection{Physiology and Doppler Waveforms}

Pulsed-wave Doppler ultrasound estimates blood-cell velocity from the frequency shift between transmitted and received waves. For target velocity $v$, $\Delta f = (2 v f_0 \cos\theta)/c_s$, where $f_0$ is transmit frequency, $\theta$ the insonation angle, and $c_s$ the speed of sound in tissue \cite{evans2011doppler}. In fetal applications, range gating samples specific vessels to produce a one-dimensional velocity envelope with sawtooth morphology: a rapid rise to peak systolic velocity (PSV), followed by decay toward end-diastolic velocity (EDV) during relaxation \cite{giles1985clinical}. EDV reflects placental vascular resistance; placental dysfunction reduces EDV, with absent end-diastolic flow indicating approximately 40\% loss of the placental vascular tree and reversed flow associated with a tenfold increase in perinatal mortality \cite{giles1985clinical}.

\subsection{ECG--Doppler Temporal Relationships and Clinical Indices}

Electromechanical coupling creates reproducible timing between ECG and Doppler features. QRS onset precedes PSV by the pre-ejection period (PEP) plus rapid ejection time; in near-term fetuses, PEP is approximately $70.0\pm2.4$\,ms and comprises electromechanical delay plus isovolumetric contraction \cite{mensahbrown2010electromechanical}. Ventricular systole ends with aortic valve closure about $225\pm13$\,ms after QRS onset, coinciding with the terminal T-wave and transition to diastole \cite{khandoker2009cardiac}.

\subsection{Generative Models for Physiological Signals}

Early physiological waveform synthesis relied on mathematical models: McSharry \textit{et al.} generated synthetic ECG via summed Gaussians \cite{mcsharry2003dynamical}, and Sameni \textit{et al.} extended this to multichannel ECG \cite{sameni2007multichannel}, though such models cannot capture the full complexity of real signals. Deep learning overcame this limitation: dilated convolution networks achieve large receptive fields for ECG synthesis \cite{oord2016wavenet, wulan2020generating}, GANs produce synthetic fetal heart rate traces \cite{zhang2022fhrgan}, and diffusion models enable multimodal conditional generation \cite{karim2025diffusets}.

\subsection{Cross-Modal Synthesis of Physiological Signals}

Cross-modal synthesis infers an inaccessible modality from an easily acquired one. PPG-to-ECG reconstruction \cite{zhu2021learning} and accelerometer-to-pulse-wave reconstruction \cite{zschocke2021reconstruction} have demonstrated feasibility via deep learning. For fetal monitoring, Verma \textit{et al.} synthesised Doppler envelopes from fECG using a U-Net \cite{verma2025reconstruction}, and Rafiei \textit{et al.} introduced Auto-FEDUS, an autoregressive dilated convolution model for fECG-to-Doppler translation \cite{rafiei2024autofedus}. Neither approach incorporated maternal ECG despite physiological evidence of maternal--fetal coupling.

\subsection{Attention Mechanisms and Multi-Channel Fusion}

Attention mechanisms assign varying weights to inputs based on learned relevance \cite{vaswani2017attention}. In maternal--fetal signal processing, attention-based architectures have been applied to fetal ECG extraction from mixed abdominal recordings \cite{wang2023correlation, qu2025integrating, nguyen2024fetal} and to fetal Doppler analysis for gestational age estimation, quality assessment, and heart rate estimation~\cite{katebi2023hierarchical, motie2025self, rafiei2025next}. These works demonstrate the value of selective processing, but to our knowledge, attention has not been applied to cross-modal synthesis of fetal Doppler envelopes from joint maternal--fetal ECG.

\begin{table*}[t]
\centering
\caption{Signal Extrapolation Performance Metrics Comparison Across Architectures. Values: mean $\pm$ SD across 5-fold cross-validation.}
\label{tab:ml_metrics}
\begin{threeparttable}
\begin{tabular}{lcccc}
\toprule
\textbf{Metric} & \makecell{\textbf{Single-Channel}\\\textit{(1-ch, no attn)}} & \makecell{\textbf{Two-Channel}\\\textit{(2-ch, no attn)}} & \makecell{\textbf{Cross-Attention}\\\textit{(2-ch, cross-attn)}} & \makecell{\textbf{Combined Attention}\\\textit{(2-ch, cross+self)}} \\
\midrule
\multicolumn{5}{l}{\textit{Pointwise error}} \\
MAE $\downarrow$ & $0.407 \pm 0.100$ & $0.403 \pm 0.082$ & $0.404 \pm 0.096$ & $\mathbf{0.403 \pm 0.103}$ \\
MSE $\downarrow$ & $0.243 \pm 0.108$ & $\mathbf{0.235 \pm 0.085}$ & $0.244 \pm 0.104$ & $0.245 \pm 0.111$ \\
RMSE $\downarrow$ & $0.481 \pm 0.102$ & $\mathbf{0.476 \pm 0.082}$ & $0.483 \pm 0.098$ & $0.482 \pm 0.107$ \\
Gradient MAE $\downarrow$ & $0.034 \pm 0.007$ & $\mathbf{0.034 \pm 0.007}$ & $0.035 \pm 0.007$ & $0.035 \pm 0.007$ \\
\midrule
\multicolumn{5}{l}{\textit{Correlation}} \\
Cross-correlation (zero lag) $\uparrow$ & $\mathbf{0.287 \pm 0.378}$ & $0.268 \pm 0.365$ & $0.279 \pm 0.364$ & $0.270 \pm 0.384$ \\
\midrule
\multicolumn{5}{l}{\textit{Sequence alignment}} \\
DTW (a.u.) $\downarrow$ & $287.2 \pm 66.9$ & $340.0 \pm 69.4$\textsuperscript{$\dagger$} & $261.8 \pm 60.8$ & $\mathbf{253.7 \pm 59.7}$ \\
\midrule
\multicolumn{5}{l}{\textit{Distribution similarity}} \\
KLD $\downarrow$ & $10.64 \pm 2.38$ & $13.17 \pm 2.08$ & $\mathbf{9.12 \pm 2.43}$ & $9.19 \pm 2.39$ \\
\midrule
\multicolumn{5}{l}{\textit{Frequency domain}} \\
PSD MSE (dB$^2$) $\downarrow$ & $84.1 \pm 17.1$ & $102.0 \pm 20.0$ & $61.8 \pm 16.6$ & $\mathbf{49.9 \pm 15.8}$ \\
PSD correlation $\uparrow$ & $0.957 \pm 0.012$ & $0.948 \pm 0.012$ & $0.962 \pm 0.011$ & $\mathbf{0.970 \pm 0.010}$ \\
\bottomrule
\end{tabular}
\begin{tablenotes}
\small
\item $\mathbf{Bold}$: Best performance. $\uparrow$: higher is better; $\downarrow$: lower is better.
\item $\dagger$ Two-channel DTW is highest despite best MSE/RMSE, indicating that minimising pointwise error yields amplitude-correct but temporally misaligned envelopes.
\end{tablenotes}
\end{threeparttable}
\end{table*}

\section{Methods}

\subsection{Dataset and Preprocessing}

We used the NInFEA dataset \cite{sulas2021ninfea}, which provides synchronized abdominal ECG and pulsed-wave Doppler recordings from 39 pregnant women between 21 and 27 weeks of gestation. The dataset was collected under approval from the Independent Ethics Committee of the Cagliari University Hospital (AOU Cagliari) following the Helsinki Declaration; all participants provided written informed consent \cite{sulas2021ninfea}. From the available recordings, we extracted 885 paired segments, each 3.75\,s long. At typical fetal heart rates (110--160\,bpm), this duration spans six to ten cardiac cycles, capturing sufficient temporal context for estimating indices such as PSV and EDV and allowing our model to learn interbeat patterns; prior work on cross-modal ECG reconstruction found that multiple consecutive beats yield more accurate generative models than single-beat inputs \cite{omer2025beatbybeat}. Each pair comprised one fetal ECG channel and the corresponding maternal ECG channel. We applied envelope extraction as described in the dataset paper \cite{sulas2021ninfea} to extract Doppler velocity envelopes, which were then low-pass filtered (4th-order Butterworth, 10~Hz cutoff) to attenuate high-frequency artifacts. All signals were resampled to 284~Hz and normalised per segment to zero mean and unit variance.



\subsection{Architecture Design}
Figure~\ref{fig:model_framework} illustrates the proposed cross-modal generative architecture, which comprises three main components: cross-modal attention for selective maternal-fetal fusion, dilated residual blocks for multi-scale temporal modeling, and self-attention for capturing long-range dependencies.

\textbf{Cross-Modal Attention}: Convolutional encoders extract features from fetal ($\mathbf{F}_f$) and maternal ($\mathbf{F}_m$) ECG. Multi-head cross-attention \cite{vaswani2017attention} with four heads computes:
\begin{equation}
\text{Attention}(\mathbf{Q}, \mathbf{K}, \mathbf{V}) = \text{softmax}\left(\frac{\mathbf{Q}\mathbf{K}^T}{\sqrt{d_k}}\right)\mathbf{V}
\end{equation}
where $\mathbf{Q} = \mathbf{F}_f\mathbf{W}^Q$, $\mathbf{K} = \mathbf{F}_m\mathbf{W}^K$, $\mathbf{V} = \mathbf{F}_m\mathbf{W}^V$, and $d_k = 64$. This selective fusion conditions maternal information on fetal features, reflecting physiological maternal-fetal coupling.

\textbf{Dilated Residual Blocks}: Eight residual blocks with dilated convolutions \cite{yu2016dilated} at rates $d \in \{2^0, \ldots, 2^7\}$ expand the receptive field to spanning multiple cardiac cycles (0.88 seconds) without significant parameter growth. Each block applies gated activation, which empirically outperforms standard ReLU activations for sequential signal modeling \cite{oord2016pixelcnn}:
\begin{equation}
\mathbf{z}_k = \tanh(\mathbf{W}_{f,k} \ast_{d_k} \mathbf{x}_k) \odot \sigma(\mathbf{W}_{g,k} \ast_{d_k} \mathbf{x}_k)
\end{equation}
where $\ast_{d_k}$ denotes dilated convolution. Outputs split into residual ($\mathbf{x}_{k+1} = \mathbf{x}_k + \mathbf{W}_{r,k}^{1 \times 1} \mathbf{z}_k$) and skip ($\mathbf{s}_k = \mathbf{W}_{s,k}^{1 \times 1} \mathbf{z}_k$) paths. Aggregated skips $\mathbf{S} = \sum_{k=1}^{8} \mathbf{s}_k$ combine multi-scale features from dilated convolution blocks to model both local pattern and global temporal structure in Doppler waveforms.

\textbf{Self-Attention}: Multi-head self-attention processes $\mathbf{S}$ with $\mathbf{Q} = \mathbf{K} = \mathbf{V} = \mathbf{S}\mathbf{W}^{QKV}$ to capture long-range temporal dependencies across cardiac cycles before dense layers produce the Doppler waveform.

\textbf{Model Variants}: To quantify the contributions of each component, we implemented four variants: (1) \emph{single-channel}, using only fetal ECG; (2) \emph{two-channel}, concatenating fetal and maternal ECG without attention; (3) \emph{cross-attention}, employing cross-modal attention but no self-attention; and (4) \emph{combined attention}, incorporating both cross-modal and self-attention.

\subsection{Loss Function Design}

We trained the model using a composite loss that encourages pointwise fidelity, smooth temporal gradients and morphological similarity:
\begin{equation}
\mathcal{L}_{\text{total}} = \mathcal{L}_{\text{MAE}} + \alpha\,\mathcal{L}_{\text{deriv}} + \alpha\,\mathcal{L}_{\text{corr}} ,
\end{equation}
where $\mathcal{L}_{\text{MAE}} = \frac{1}{T}\sum_t |\hat{y}_t - y_t|$ is the mean absolute error, $\mathcal{L}_{\text{deriv}} = \frac{1}{T}\sum_t |(\hat{y}_t - \hat{y}_{t-1}) - (y_t - y_{t-1})|$ penalises differences in first derivatives to promote physiologically realistic systolic upstrokes and diastolic decay, and $\mathcal{L}_{\text{corr}} = 1 - \rho(\hat{y},y)$ encourages high Pearson correlation between generated and real envelopes. We set $\alpha = 0.5$ to assign equal weight to each auxiliary objective, ensuring approximately balanced gradient contributions from pointwise error, derivative matching, and correlation alignment. This choice was informed by monitoring individual loss component magnitudes during preliminary training runs, which confirmed that equal weighting prevented any single term from dominating the total gradient and yielded stable convergence.

\subsection{Implementation Details}

Models were trained using the Adam optimiser with an initial learning rate of $10^{-3}$. The learning rate was reduced by a factor of 0.5 if the validation loss plateaued for five epochs, with a minimum of $10^{-5}$. Early stopping with a patience of 20 epochs prevented overfitting. We used a batch size of 8 and trained for up to 100 epochs per fold. Fivefold cross-validation with patient-level splitting assessed generalisation; no patient overlapped between training and validation sets.

\subsection{Evaluation Metrics}

Reconstruction performance was assessed using metrics that capture time-domain, frequency-domain, and distributional properties. Pointwise errors included the mean absolute error (MAE), mean squared error (MSE), root mean squared error (RMSE), and gradient MAE (error in first derivatives). Correlation measures comprised the Pearson correlation coefficient and zero-lag cross-correlation. Temporal alignment was quantified using dynamic time warping (DTW) distance; since all signals are normalised to zero mean and unit variance prior to evaluation, DTW values are in normalised amplitude units (a.u.). Distributional similarity was measured by the Kullback--Leibler divergence (KLD) between histograms of generated and real envelopes. Frequency-domain fidelity was evaluated by the MSE and correlation of the power spectral density (PSD). PSD was estimated using Welch's method~\cite{welch1967use} with a sampling rate of $f_s = 284$\,Hz and a segment length of $n_{\text{seg}} = 256$ samples, yielding a frequency resolution of approximately 1.1\,Hz. PSD values were converted to decibels ($10\log_{10}(\text{PSD} + \epsilon)$, where $\epsilon = 10^{-10}$) before computing the MSE, so PSD MSE is reported in units of dB$^2$. Clinical relevance was assessed by computing MAE and correlation for PI, RI, S/D ratio, PSV, EDV, time-averaged maximum (TAMX), and heart rate.

\section{Results}

\subsection{Overall Performance}

Table~\ref{tab:ml_metrics} summarizes reconstruction performance. The combined attention model achieves the lowest DTW ($253.7\pm59.7$\,a.u.) and best frequency-domain fidelity: PSD MSE of $49.9\pm15.8$\,dB$^2$, which is 51\% lower than the two-channel baseline, and PSD correlation of $0.970\pm0.010$. Na\"ively concatenating maternal and fetal ECG yields the lowest MSE and RMSE but the worst DTW, indicating that pointwise accuracy alone does not guarantee morphological alignment. By contrast, cross-modal and self-attention improve both temporal alignment and frequency-domain fidelity.

\subsection{Ablation Analysis}

\textbf{Maternal ECG paradox}: Single-channel and two-channel baselines yield nearly identical pointwise error (MAE 0.407 vs. 0.403) and correlation, while the two-channel variant has a substantially worse DTW. This shows that na\"ive inclusion of maternal ECG provides no benefit and can degrade temporal structure.

\textbf{Cross-attention impact}: Introducing cross-channel attention dramatically improves performance relative to the na\"ive two-channel model. As shown in Table~\ref{tab:ml_metrics}, cross-attention reduces DTW by 23\%, KLD by 31\% and PSD MSE by 39\% compared with simple concatenation, confirming that selective incorporation of maternal information is critical for capturing maternal--fetal coupling.

\textbf{Self-attention contribution}: Adding self-attention on top of cross-attention yields further gains. Self-attention provides an additional 19\% improvement in PSD MSE (from $61.8\pm16.6$ to $49.9\pm15.8$\,dB$^2$) and a modest 3\% improvement in DTW, reflecting its ability to capture long-range temporal dependencies and beat-to-beat consistency.

\begin{figure}[t]
\centering
\includegraphics[width=\linewidth]{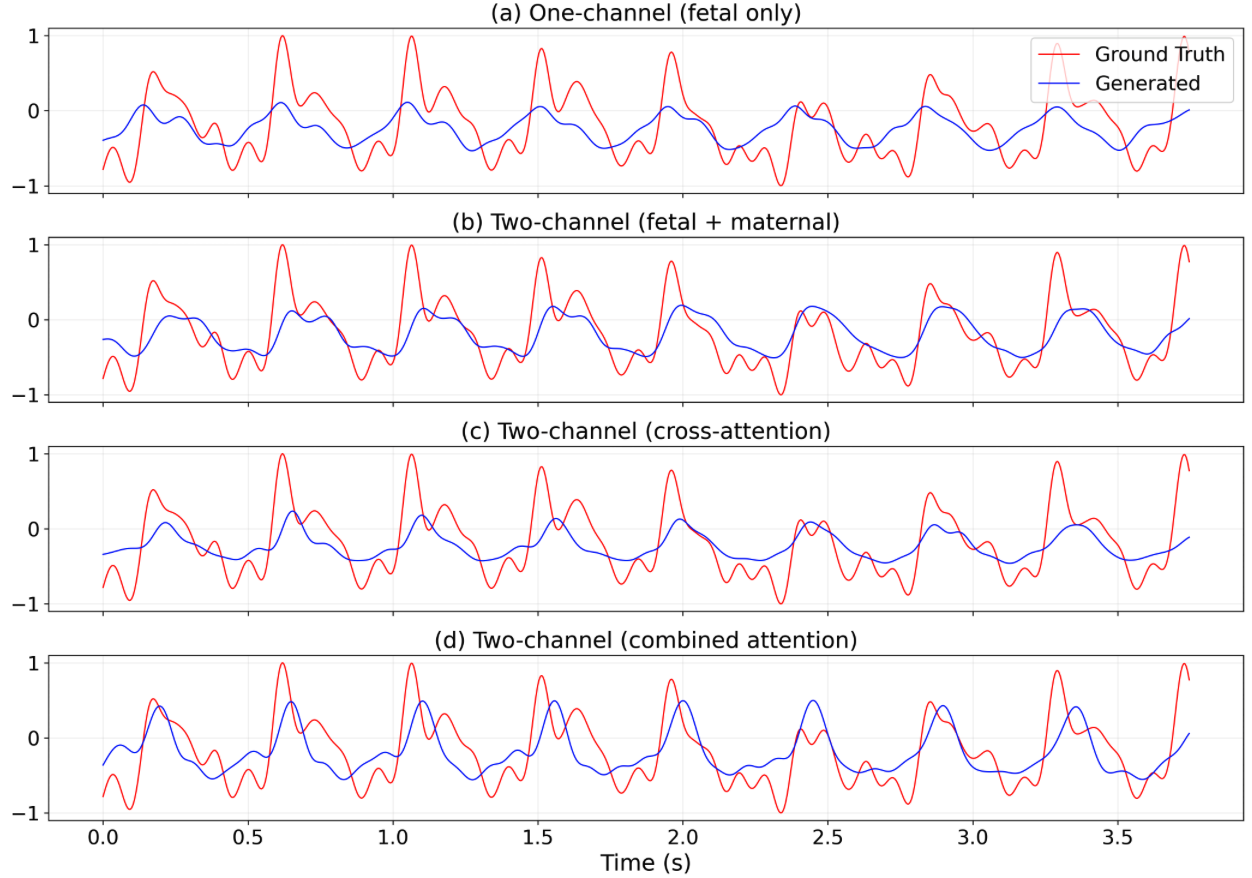}
\caption{Example Doppler velocity envelopes generated by each model variant. Each panel shows the generated envelope (blue) superimposed on the ground-truth envelope (red): (a)~single-channel model (fetal ECG only), (b)~two-channel model (fetal and maternal ECG) without attention, (c)~two-channel cross-attention model, and (d)~two-channel combined attention model. The x-axis is time in seconds (at 284\,Hz sampling rate; 1065 samples $\approx$ 3.75\,s); the y-axis shows normalised Doppler amplitude (zero mean, unit variance). Progressive improvements in temporal alignment and morphological fidelity are visible from (a) to (d): note how the systolic peaks in (d) align more closely with the ground truth than in (a) or (b).}
\label{fig:example_variants}
\end{figure}

Example Doppler velocity envelopes generated by each model variant are shown in Figure~\ref{fig:example_variants}. Each panel compares the generated envelope (blue) with the ground truth (red) for a representative segment, illustrating how attention mechanisms progressively improve temporal alignment and morphological fidelity.

\subsection{Clinical Metrics Performance}

\begin{table*}[t]
\centering
\caption{Clinical/Physiological Performance Metrics Comparison Across Architectures. For each metric, we report the MAE between real and generated values and the Pearson correlation. Values reported as mean $\pm$ SD across 5-fold cross-validation.}
\label{tab:clinical_metrics}
\begin{threeparttable}
\resizebox{\textwidth}{!}{%
\begin{tabular}{lcccccccc}
\toprule
& \multicolumn{2}{c}{\makecell{\textbf{Single-Channel}\\\textit{(1-ch, no attn)}}} & \multicolumn{2}{c}{\makecell{\textbf{Two-Channel}\\\textit{(2-ch, no attn)}}} & \multicolumn{2}{c}{\makecell{\textbf{Cross-Attention}\\\textit{(2-ch, cross-attn)}}} & \multicolumn{2}{c}{\makecell{\textbf{Combined Attention}\\\textit{(2-ch, cross+self)}}} \\
\cmidrule(lr){2-3} \cmidrule(lr){4-5} \cmidrule(lr){6-7} \cmidrule(lr){8-9}
\textbf{Metric} & MAE $\downarrow$ & Corr. $\uparrow$ & MAE $\downarrow$ & Corr. $\uparrow$ & MAE $\downarrow$ & Corr. $\uparrow$ & MAE $\downarrow$ & Corr. $\uparrow$ \\
\midrule
\multicolumn{9}{l}{\textit{Hemodynamic indices}} \\
PI & $48.91 \pm 40.09$ & $-0.02 \pm 0.09$ & $49.40 \pm 39.13$ & $-0.02 \pm 0.08$ & $49.11 \pm 39.65$ & $-0.08 \pm 0.13$ & $\mathbf{48.78 \pm 39.36}$ & $-0.04 \pm 0.08$ \\
RI & $\mathbf{19.11 \pm 28.28}$ & $\mathbf{0.07 \pm 0.05}$ & $26.39 \pm 31.06$ & $0.00 \pm 0.03$ & $91.99 \pm 106.68$ & $-0.04 \pm 0.09$ & $30.41 \pm 40.56$ & $0.06 \pm 0.06$ \\
S/D ratio & $0.74 \pm 0.33$ & $-0.02 \pm 0.06$ & $0.92 \pm 0.27$ & $-0.01 \pm 0.10$ & $0.72 \pm 0.15$ & $-0.02 \pm 0.05$ & $\mathbf{0.66 \pm 0.19}$ & $0.04 \pm 0.09$ \\
\midrule
\multicolumn{9}{l}{\textit{Cardiac function}} \\
Heart rate (bpm) & $4.78 \pm 0.69$ & $0.58 \pm 0.15$ & $4.80 \pm 1.13$ & $0.55 \pm 0.20$ & $\mathbf{4.69 \pm 0.71}$ & $\mathbf{0.58 \pm 0.14}$ & $4.71 \pm 0.77$ & $0.57 \pm 0.12$ \\
\midrule
\multicolumn{9}{l}{\textit{Velocity components}} \\
PSV & $0.61 \pm 0.15$ & $0.09 \pm 0.17$ & $0.71 \pm 0.11$ & $0.13 \pm 0.17$ & $0.64 \pm 0.11$ & $-0.06 \pm 0.13$ & $\mathbf{0.60 \pm 0.12}$ & $0.03 \pm 0.14$ \\
EDV & $0.41 \pm 0.18$ & $0.15 \pm 0.10$ & $0.52 \pm 0.19$\textsuperscript{$\ddagger$} & $0.11 \pm 0.10$ & $\mathbf{0.33 \pm 0.08}$ & $0.03 \pm 0.08$ & $0.34 \pm 0.09$ & $-0.04 \pm 0.10$ \\
TAMX & $0.12 \pm 0.02$ & $0.00 \pm 0.05$ & $0.12 \pm 0.01$ & $-0.02 \pm 0.14$ & $0.12 \pm 0.01$ & $-0.07 \pm 0.15$ & $\mathbf{0.12 \pm 0.02}$ & $-0.04 \pm 0.17$ \\
\bottomrule
\end{tabular}}
\begin{tablenotes}
\small
\item $\mathbf{Bold}$: Best performance (lowest MAE or highest correlation). $\uparrow$: higher is better; $\downarrow$: lower is better.
\item $\ddagger$ The two-channel model yields higher EDV error due to misaligned diastolic morphology, consistent with its poor DTW alignment.
\item Legend: PI: pulsatility index; RI: resistance index; S/D: systolic/diastolic ratio; PSV: peak systolic velocity; EDV: end diastolic velocity; \\TAMX: time-averaged maximum velocity.
\end{tablenotes}
\end{threeparttable}
\end{table*}

As summarized in Table~\ref{tab:clinical_metrics}, all models achieved clinically acceptable heart rate accuracy (MAE 4.69--4.80\,bpm; correlation 0.55--0.58). Given that fetal bradycardia and tachycardia are defined as heart rates below 110\,bpm and above 160\,bpm, respectively \cite{bhide2013isuog}, an absolute error of less than 5\,bpm is negligible relative to the physiological range. Combined attention achieved the best performance in restoring S/D ratio (MAE 0.66), which is crucial for detecting elevated placental resistance.

\section{Discussion}

\subsection{Recoverable and Residual Components of Doppler Waveforms}

Our framework decomposes fetal Doppler into components recoverable from ECG and residual components that cannot be reconstructed. The strong single-channel performance demonstrates that fetal ECG encodes substantial Doppler information through excitation-contraction coupling \cite{bers2002cardiac}: R--R intervals determine pulsation frequency, and QRS timing predicts systolic onset via the pre-ejection period \cite{mensahbrown2010electromechanical}. These electrically recoverable components confirm that electromechanical coupling provides a robust basis for cross-modal synthesis.

The pattern of metric improvements reveals which Doppler features depend on electrical versus mechanical factors. Attention mechanisms yielded substantial gains in frequency-domain fidelity (PSD MSE: $49.9$ vs.\ $84.1$\,dB$^2$ for single-channel, 41\% improvement) and temporal alignment (DTW: $253.7$ vs.\ $287.2$\,a.u., 12\% improvement), indicating that overall waveform morphology and timing are electrically recoverable when maternal coupling is appropriately modeled. Heart rate estimation also benefited modestly ($4.71$ vs.\ $4.78$\,bpm error), consistent with its direct dependence on R--R intervals.

However, clinical indices derived from specific velocity components showed limited or no improvement. PI and RI correlations remained near zero across all architectures (PI: $r = -0.04$; RI: $r = 0.06$), and velocity metrics (PSV, EDV, TAMX) showed weak correlations despite reasonable MAE values. This dissociation between good waveform reconstruction and poor index correlation suggests that these indices depend on beat-to-beat mechanical variations, including ventricular compliance, placental resistance, and preload, that are invisible to electrical recordings \cite{hecher1995fetal, baschat2004doppler}. The model learns average morphology well but cannot capture the mechanical factors driving inter-beat variability in peak and end-diastolic velocities. These residual components may be clinically significant precisely because they reflect mechanical properties inaccessible to fECG monitoring.

\subsection{Selective Attention for Maternal--Fetal Cardiac Coupling}

The finding that na\"ive maternal ECG concatenation provides zero benefit, and degrades DTW alignment by 18\%, parallels physiological evidence that MFCC occurs intermittently \cite{vanleeuwenpnas2009}. When coupling is weak, maternal signals contribute noise. Cross-modal attention addresses this by learning to emphasize temporally correlated maternal segments while suppressing irrelevant portions.

The 39\% PSD MSE improvement from cross-modal attention over na\"ive concatenation (from $101.9$ to $61.8$\,dB$^2$) quantifies the value of learned selectivity. This aligns with transfer entropy analyses showing that maternal-to-fetal information transfer varies with gestational age and fetal behavioral state \cite{marzbanrad2015transfer}. By treating fetal features as queries and maternal features as keys/values, cross-modal attention provides an architectural analogue to this conditional coupling. Self-attention contributes an additional 19\% frequency-domain improvement (from $61.8$ to $49.9$\,dB$^2$) by capturing long-range temporal dependencies within the fetal signal.

Notably, the benefits of attention were domain-specific: frequency-domain and alignment metrics improved substantially, while pointwise error metrics (MAE, MSE) showed minimal differences across architectures. This dissociation arises because all architectures learn similar average waveform morphologies, yielding comparable pointwise error, whereas frequency-domain metrics and DTW capture spectral fidelity and temporal alignment of cardiac events, properties that depend on modeling the \textit{timing} of systolic and diastolic phases across beats. Cross-modal attention improves these timing-sensitive metrics by selectively incorporating maternal heart rate information that modulates beat timing, while self-attention enforces beat-to-beat temporal consistency. This pattern is consistent with the physiological expectation that electromechanical coupling primarily determines \textit{when} cardiac events occur, while local amplitude depends on mechanical properties such as contractility and vascular impedance.

\subsection{Comparison to Prior Work}

Existing fetal Doppler synthesis methods use fetal ECG exclusively \cite{verma2025reconstruction, rafiei2024autofedus}. Our work is the first to incorporate maternal ECG through attention mechanisms, explicitly modeling MFCC. Beyond improved reconstruction, this framework enables new scientific questions: What proportion of Doppler variability reflects maternal--fetal interactions? Which features require direct hemodynamic measurement? The attention-based decomposition provides quantitative answers: cross-modal attention contributes 39\% of frequency-domain improvement, while residual errors identify features inaccessible to electrical recordings.

\subsection{Limitations and Future Directions}

Our dataset comprises 39 mid-gestation, predominantly healthy pregnancies. Since maternal--fetal cardiac coupling patterns may change significantly under pathological conditions such as fetal growth restriction or pre-eclampsia, where altered autonomic regulation and increased placental resistance fundamentally change ECG--Doppler relationships \cite{khandoker2019mfcc}, validation on pathological cohorts is essential. The persistent difficulty in reconstructing clinical indices suggests fundamental limits to ECG-based Doppler synthesis for metrics dependent on mechanical factors. Additionally, the reported improvements are based on descriptive comparison of mean $\pm$ SD across five folds; formal per-segment statistical testing would further strengthen the evidence for architectural differences, particularly for metrics where improvements are modest.

From a clinical perspective, this framework has potential applications in continuous fetal monitoring where Doppler ultrasound is impractical. Since fECG can be acquired continuously via wearable electrodes while Doppler requires episodic assessment by trained personnel, fECG-derived envelopes could enable real-time heart rate and waveform morphology estimation during ambulatory monitoring. However, the limited correlation in clinical indices indicates that synthesized waveforms should not replace direct Doppler for detecting placental dysfunction. Future work should investigate physiological constraints (e.g., one PSV per R-peak), uncertainty quantification for clinical deployment, whether residual components correlate with mechanical abnormalities detectable only through direct Doppler measurement, and morphological qualitative assessment of generated waveforms by obstetrics and ultrasound clinicians to complement automated metrics.

\section{Conclusion}

We presented a cross-modal generative framework for investigating relationships between fetal Doppler and fetal-maternal ECG. Critical ablation studies established that na\"ive maternal ECG concatenation provides no benefit, while selective attention yields 39\% frequency-domain improvement, validating that MFCC requires learned selectivity for computational exploitation. The combined attention model achieved high-fidelity fetal Doppler reconstruction in the frequency domain, indicated by DTW of $253.7$\,a.u., PSD MSE of $49.9$\,dB$^2$, PSD correlation of $0.970$, and heart rate error of $4.71$\,bpm. However, the differential response to attention mechanisms, with substantial gains in waveform morphology but limited improvement in clinical indices (PI, RI), reveals the boundary between electrically recoverable and mechanically determined Doppler components. In particular, hemodynamic indices such as PI and RI showed near-zero correlations across all architectures, indicating a fundamental limitation of ECG-based Doppler synthesis for metrics dependent on beat-to-beat mechanical variability. By delineating which aspects of fetal hemodynamics are accessible through non-invasive fECG and which remain exclusive to direct Doppler measurement, our work provides a foundation for future fetal monitoring research.

\end{document}